\pdfoutput=1
\documentclass[sigconf]{aamas} 

\usepackage[subtle]{savetrees}
\definecolor{forestgreen}{rgb}{0., 0.5, 0.} 
\usepackage{enumitem}
\usepackage{ragged2e}
\usepackage{makecell, multirow}
\usepackage{tabularx}
\newcolumntype{L}{>{\RaggedRight}X}
\usepackage{balance} 
\usepackage{filecontents,catchfile}
\usepackage[english]{babel}
\usepackage[utf8]{inputenc}
\usepackage{algorithm}
\usepackage[noend]{algpseudocode}
\usepackage{siunitx}

\setcopyright{ifaamas}
\acmConference[AAMAS '23]{Proc.\@ of the 22nd International Conference
on Autonomous Agents and Multiagent Systems (AAMAS 2023)}{May 29 -- June 2, 2023}
{London, United Kingdom}{A.~Ricci, W.~Yeoh, N.~Agmon, B.~An (eds.)}
\copyrightyear{2023}
\acmYear{2023}
\acmDOI{}
\acmPrice{}
\acmISBN{}

\title[AAMAS-2023 Formatting Instructions]{Joint Engagement Classification using Video Augmentation Techniques for Multi-person Human-robot Interaction}

\author{Yubin Kim, Huili Chen, Sharifa Alghowinem, Cynthia Breazeal, and Hae Won Park}
\affiliation{
  \institution{MIT Media Lab}
  \city{Cambridge}
  \state{Massachusetts}}

\begin{abstract}
Affect understanding capability is essential for social robots to autonomously interact with a group of users in an intuitive and reciprocal way. However, the challenge of multi-person affect understanding comes from not only the accurate perception of each user's affective state (e.g., engagement) but also the recognition of the affect interplay between the members (e.g., joint engagement) that presents as complex, but subtle, nonverbal exchanges between them. Here we present a novel hybrid framework for identifying a parent-child dyad's joint engagement by combining a deep learning framework with various video augmentation techniques. Using a dataset of parent-child dyads reading storybooks together with a social robot at home, we first train RGB frame- and skeleton-based joint engagement recognition models with four video augmentation techniques (General Aug, DeepFake, CutOut, and Mixed) applied datasets to improve joint engagement classification performance. Second, we demonstrate experimental results on the use of trained models in the robot-parent-child interaction context. Third, we introduce a behavior-based metric for evaluating the learned representation of the models to investigate the model interpretability when recognizing joint engagement. This work serves as the first step toward fully unlocking the potential of end-to-end video understanding models pre-trained on large public datasets and augmented with data augmentation and visualization techniques for affect recognition in the multi-person human-robot interaction in the wild.

\end{abstract}


\newcommand{\BibTeX}{\rm B\kern-.05em{\sc i\kern-.025em b}\kern-.08em\TeX}

\begin{document}


\pagestyle{fancy}
\fancyhead{}


\makeatletter
\def\@ACM@checkaffil{
    \if@ACM@instpresent\else
    \ClassWarningNoLine{\@classname}{No institution present for an affiliation}%
    \fi
    \if@ACM@citypresent\else
    \ClassWarningNoLine{\@classname}{No city present for an affiliation}%
    \fi
    \if@ACM@countrypresent\else
        \ClassWarningNoLine{\@classname}{No country present for an affiliation}%
    \fi
}
\makeatother

\maketitle 

\section{Introduction}
Affective communication is essential for human-human interaction~\cite{Picard2003-AffectiveCC} and has strong links to learning~\cite{Picard2004-AffectiveLM}, persuasion~\cite{Haddock2008-ShouldPA}, and a variety of other functions. The ability of a socially interactive robot to perceive human nonverbal cues, social signals, and emotions is critical for engaging with humans in an intuitive, natural, and reciprocal manner~\cite{CHEN2020TLC,Spaulding2018-ASRS}. This affect understanding capacity has been identified as a basic robot capability required for higher-level competencies in human-robot interactions~\cite{Thomaz2016ComputationalHI} and contributes to a robot's user profiling and behavior adaptability capabilities~\cite{Rossi2017UserPB}. Consequently, the objective of our research is not only to improve the affective recognition and understanding capabilities of social robots in multi-person human-robot interaction (HRI) scenarios but also to assess the quality of the model for deployment.

Given the impact of parent-child affective exchanges in children's development and growth, parent-child interaction will be the primary application domain for our research. High-quality, reciprocal relationships between parents and children promote children's social, emotional, cognitive, and linguistic development~\cite{Tamis2001-MRA}. However, not all children have equal access to socially and emotionally rich dialogic interactions, such as being prompted by open-ended questions and back-and-forth conversations~\cite{Rowe2008-CDS}. Social robots have a compelling potential to facilitate human-human connection and conversation \cite{jung2017affective}, and they can be designed to mediate parent-child dialogic interactions by encouraging and demonstrating best pedagogical practices. In order to reach such capacity, \textbf{affect understanding in a multi-person interaction context} should support  robots to comprehend the interpersonal dynamics between a parent and a child and provide timely and appropriate actions to mediate the flow of the interaction and maintain the engagement of the dyad.

The nonverbal cues people display in a group interaction present excellent data sources for the development of contact-free, unobtrusive affect recognition systems that can comprehend human-human affective dynamics. In parent-child interactions, the parent and child's nonverbal behaviors, such as head/body movement, gestures, and postures, are particularly important indicators to gauge interaction quality including synchronization, engagement,  attachment, and shared affect~\cite{colegrove2017review,grebelsky2014parental,leclere2016interaction}. 
However, automatic perception of these affective dynamics between the members of a group via  nonverbal indicators is still much underexplored.

In an effort to further this field of study, we propose a novel hybrid method for identifying parent-child joint engagement by combining a deep learning framework with a range of video augmentation techniques -- General Aug \cite{papakipos2022augly}, DeepFake \cite{DBLP:conf/mm/ChenCNG20}, CutOut \cite{devries2017improved}, and Mixed \cite{yun2019cutmix}). Using the state-of-the-art action recognition algorithms such as SlowFast \cite{feichtenhofer2019slowfast} as base models, we applied video augmentation techniques to improve the pre-trained model's representation learning in order to raise its sensitivity to capture the subtle social cues that leads to understanding the joint engagement states of parent-child dyads. Even though SlowFast and other base models have been largely used to detect human activity and motion in prior works \cite{kim2020spatio, wang2019i3d, zhang2021current}, we show that they  also produce high performance in social cue understanding tasks with fine-tuning using video augmentation techniques. 

Using our novel evaluation metric, we also demonstrate the interpretability of the model's learned representation. The model interpretability is particularly crucial for humans to understand how the model learns the continuous dynamics of human social cues. In order to visualize where our model attends to in the video frames to identify joint engagement, we compare Gradient-weighted Class Activation Mapping (Grad-CAM) results to the social cues humans attend to gauge joint engagement, i.e., annotation guidelines given to human annotators. 

In summary, our work contributes the following :
\begin{enumerate}[leftmargin=1.5\parindent]
    \item Adapt end-to-end and skeleton-based deep learning models to joint engagement recognition task in multi-person HRI setting. The models are pre-trained on large public datasets of human action and activity recognition task and fine-tuned with video augmentation techniques for recognition of psychological processes, i.e., joint engagement, that involves changes in subtle social cues;
    \item Conduct a comprehensive analysis with experimental results to demonstrate the use of pre-trained models in an affective communication context, and provide a competitive affect recognition baseline for future multi-person affect understanding;
    \item Implement a new metric for evaluating the learned representation that compares human annotation guidelines and the visual regions the model attends to.
\end{enumerate}

\section{Related Works}
\subsection{Social-Affective cue understanding in Human-Robot Interaction (HRI)}
Developing affect signal perception systems for social robots has been extensively investigated and shown to have empirical significance, particularly in early childhood development. Social robots with this affect understanding capability have been found to promote children's learning more effectively than those without, e.g.,~\cite{Gordon2016-APS,Park2019-AMF}. Prior works have integrated user social and affective signals into either its behavior policy or cognitive model as human feedback on the robot's newly executed action to deliver real-time personalized interaction~\cite{Gordon2016-APS,Park2019-AMF}. Similarly, affective signals have been incorporated into the robot's user cognitive and skill estimation models to improve user model accuracy, such as a student vocabulary acquisition~\cite{Spaulding2016AffectSM}. Overall, affect recognition enhances the effectiveness of robots to provide timely interventions to individual users and improve their interaction experience. Nonetheless, the majority of affect-ware social robots were only designed for one-on-one interactions with humans. When interacting with a single person, it is sufficient for an intelligent system to recognize the social and affective cues directed at the interaction task or the robot. In contrast, when engaging in dyadic human interaction, interactive technology must be able to recognize the affective and social dynamics between the two users. 

To date, the vast majority of current affect recognition models, particularly commercial affect extraction tools, are only applicable in single-person settings. Only a handful of previous Multi-person HRI field studies, e.g., ~\cite{vazquez2017TRA,Strohkorb2015CCS}, investigated how to equip a robot with a perception system to recognize the social-affective dynamics of a human group. The perception system in ~\cite{vazquez2017TRA}, for instance, estimates user positions and body orientations using a Kinect to track participants and control the orientation and gaze of the robot. In a separate study~\cite{Strohkorb2015CCS}, a prediction model for a participant's social dominance in a group human-robot interaction was developed but trained on the nonverbal behavioral features that were manually handcrafted by human coders offline, e.g., utterance type, gaze, interruptions. One previous study~\cite{Salam2017-FullyAA} focused on the automatic analysis and classification of engagement based on humans' and robots' personality profiles using a dataset gathered in the context of triadic human-human-robot interaction.

Due to this limitation in the robot's perception system, the majority of current multi-party HRI research employs wizard-of-oz paradigms to teleoperate a robot or an oversimplified behavior policy (e.g., simple rule-based or tablet-based behavior triggers) that does not depend on the robot's affective perceptual capacity. In a minority of studies, machine learning or reinforcement learning was used to train the robot's behavior policy~\cite{Vazquez2016MAF,Keizer2014MLS}. Even fewer studies have equipped robots with affective perception that can guide their interactive behavior. Research on developing affect recognition models for dyadic human groups would unlock the potential for a robot to engage in and even enhance inherently complex human-human interactions. Therefore, the advancement of multi-person affect recognition would catalyze the development of fully autonomous social robots in multi-person HRI settings.

\subsection{Deep Learning Approach to Affect \\ Recognition}

The majority of affect recognition models, and commercial affect extraction tools, in particular, are primarily concerned with single-person or single-modality affect detection. In recent years, deep learning has been used extensively to develop affect detection models trained on human behavioral cues in audiovisual recordings, such as facial expression, speaking style, speech prosody, linguistics sentiment, and head and body movement (see the review~\cite{feng2020review}). Deep belief networks (DBNs)~\cite{zhang2017learning}, attentively-coupled long-short term memory (ACLSTM)~\cite{hsu2020attentively}, and multitask LSTM augmented with two-stream auto-encoder for deep feature extraction~\cite{hao2020visual} are examples of deep learning techniques used in previous research. The audio-video input features both handcrafted and deep features have been used as model input for recognition (e.g.,~\cite{hao2020visual}), while the predicted affective states range from valence and arousal to engagement (e.g.,~\cite{rudovic2018personalized}).

In contrast to the widespread application of deep learning to the detection of individual affect, affect detection in multi-person interactions is significantly less studied. Using deep learning models, a small number of studies have investigated dyadic dynamics, primarily from a single modality, e.g.,~\cite{heimerl2020transparent,Huili2020Dyadic}. A number of multi-person interaction perception models focus on human action recognition such as handshakes, e.g.\cite{weng2021human}. For instance, deep features from the full body and body parts of both individuals, as well as handcrafted motion and posture features, were extracted to train deep learning models for action recognition models, such as \cite{weng2021human}. Using a combination of deep features, a graph network, and a logic-aware module, the relationship and interaction between two individuals in still images were analyzed in \cite{wang2020lagnet}. Personality recognition in the dyadic interaction context was also modeled using the transformer-based method that utilizes multimodal deep features extracted and individuals' socio-demographic profiles (e.g. gender, relationship status, mood)~\cite{palmero2020context}. In a previous study, \citet{zhang2020multi} developed a temporal fusion of multimodal features extracted from vision, audio and text to recongize each participant's social role in a four-person meeting. 

To the best of our knowledge, only a handful of recent works have begun to develop deep models for recognizing affect in multi-person interactions (e.g., ~\cite{Chen2020DyadicSA,Li2021-ImprovingMS}). For instance, Chen and colleagues~\cite{Chen2020DyadicSA} recently utilized end-to-end deep learning methods augmented with attention mechanisms to identify each individual's affective expression in an audio stream containing the utterances of two speakers. Another work by \cite{heimerl2020transparent} developed a framework to identify an individual's engagement in the context of a two-way conversation. In their study, a hybrid approach of deep models and Bayesian networks was used to predict interpersonal dynamics in the dyadic interaction, including back channeling, speaking turns, gender, and face, hand movement, speech, and context data. The audiovisual recordings were captured separately for each interlocutor. 

Overall, very little research has been conducted on the application of a deep learning approach to multi-person affect recognition in which all the interlocutors interact with one another in a single audiovisual recording, limiting the development of affect-aware robots suited for dyadic human interactions in the real world. In addition, advanced techniques in deep learning research, such as video augmentation techniques and explainable visualization in deep learning, have been applied to advance multi-person affect perception models for social robots. Consequently, our work proposes a novel framework that employs these cutting-edge deep learning techniques to significantly enhance a robot's ability to comprehend the affective dynamics of multi-person HRI in the wild.

\subsection{Video Classification Models and Data \\ Augmentation Techniques}
State-of-the-art video classification models were mostly developed for action recognition, a central task in video understanding \cite{feichtenhofer2019slowfast, feichtenhofer2020x3d, wang2019i3d}. Among various modalities (e.g., RGB frames, optical flow, human skeleton, and audio waves) used for feature representation, RGB-based and Skeleton-based action recognition models have been the mainstream approach in recent years \cite{wang2019i3d, feichtenhofer2019slowfast, feichtenhofer2020x3d, bertasius2021spacetime}. RGB frames are the most basic and typical modality to be used for model training in action recognition tasks. The recently proposed TimeSformer \cite{bertasius2021spacetime} network is only built on self-attention over space and time. It adapts the Transformer architecture to video by enabling spatiotemporal feature learning directly from a sequence of frame-level patches. X3D \cite{feichtenhofer2020x3d} is a family of efficient video networks that continuously expand a small 2D image classification architecture along multiple network axes (space, time, width, and depth). I3D \cite{wang2019i3d} is a 2D ConvNet inflation-based model, in which the filters and pooling kernels of deep image classification ConvNets are expanded into 3D. In SlowFast \cite{feichtenhofer2019slowfast}, a dual-pathway structure is proposed to combine the benefits of a slow pathway for static spatial features and a fast pathway for dynamic motion features. 

On the other hand, human skeletons in a video provide a sequence of joint coordinate lists which emphasizes its action-focusing nature and compactness. Recently proposed Channel-wise Topology Refinement Graph Convolution Network (CTR-GCN) \cite{chen2021channel} dynamically learns different topologies and effectively aggregates joint features in different channels. The multi-Scale aggregation Scheme (MS-G3D) \cite{liu2020disentangling} disentangles the importance of nodes in different neighborhoods for effective long-range modeling. It utilizes dense cross-spacetime edges as skip connections for direct information propagation across the spatial-temporal graph. STGCN \cite{duan2022PYSKL} adopts Graph Convolution Neural (GCN) Networks for skeleton processing. Based on all the above works, Transformer, CNN, and Graph Convolutional models have achieved breakthrough results in action recognition tasks. With the great power of understanding general human activities, pre-trained action recognition models make it suitable for the models to solve the task of joint engagement recognition with the fine-tuning process. 



\begin{figure*}
  \includegraphics[width=1.0\linewidth]{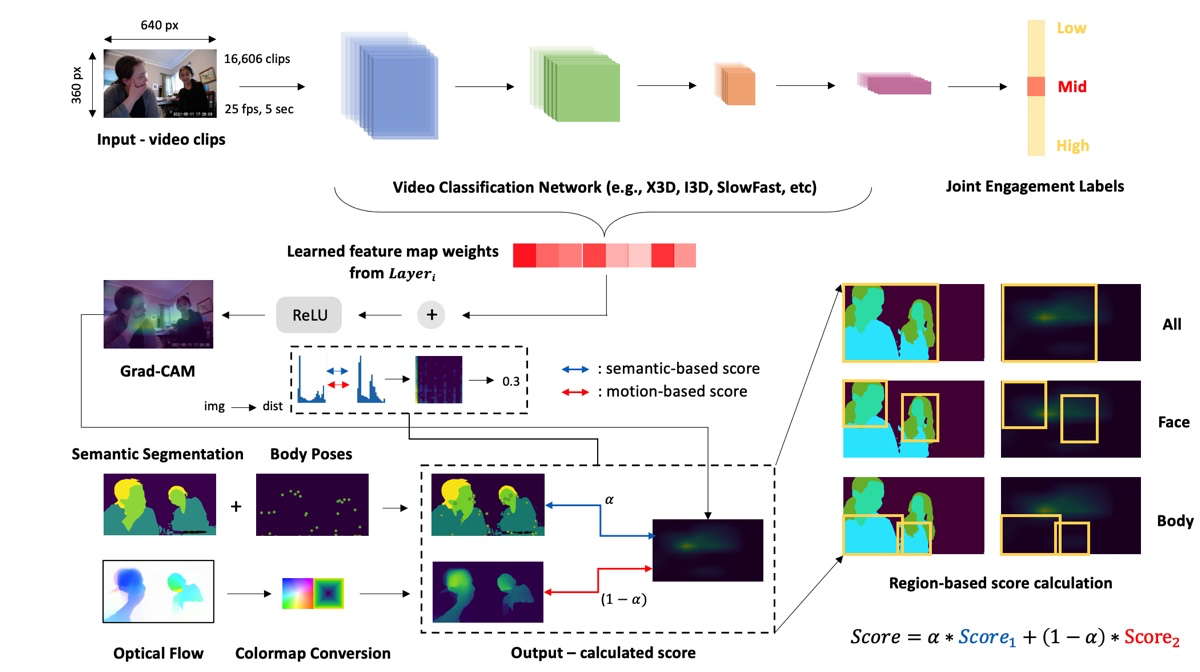}
  \caption{Proposed framework for evaluating the learned representation (Grad-CAM) with modified Optical Flow and skeleton information combined Semantic Segmentation as references. With the fine-tuned models, we generate Grad-CAM for each video clip and evaluate its quality. We calculate the evaluation score based on two sub-scores (semantic-based and motion-based) which are obtained by applying mutual information and cross-entropy. }
  \label{fig:framework}
\end{figure*}


\section{Methods}

In this section, we introduce the robot-parent-child interaction dataset our models were trained on, and four different video augmentation methods. The behavior-based objective metric to evaluate the learned representation is also presented. 

\subsection{Dataset and Annotation}
\label{sec:dataset}
The dataset we used to train our affect model originated from our previous deployment~\footnote{dataset publication under review at the time of submission of this paper to AAMAS2023}. In the study, a social robot was deployed and teleoperated remotely in the homes of 12 families with 3-7-year-old children to engage in a triadic story-reading activity with the parent and child over the course of six 25-minute sessions and 3 to 6 weeks in total. For each triadic session, audiovisual recordings were captured and subsequently used to annotate the quality of parent-child engagement. We chose the Joint Engagement Rating Inventory (JERI) to measure parent-child engagement~\cite{Adamson2018}, as it has been utilized and validated in previous parent-child interaction studies~\cite{Adamson2018,Chen2022-DyadicAP}. This engagement metric quantifies and classifies both the verbal and nonverbal behaviors associated with a child's interaction with its parent. 

To annotate the parent-child joint engagement in the audio-visual recordings of the parent-child-robot interactions, we recruited two trained annotators with a psychology or education background. The coding scheme, the choice of the video interval threshold, and the annotation protocol were derived from previous work on joint engagement in parent-child interaction~\cite{Chen2022-DyadicAP}. Specifically, the annotators gave ratings every five non-overlapping seconds on a five-point ordinal scale [-2,2], with two corresponding to cases in which the parent-child pair displayed clear signs of high joint engagement valence and -2 corresponding to cases in which the parent-child pair displayed clear signs of low joint engagement valence. The video fragment interval of target audio-visual recordings used to generate continuous quality scales was determined to be five seconds. 

Using the intra-class correlation (ICC) type (3,1) for average fixed raters, the agreement among the three annotators was measured. Given these evaluation criteria, the annotation quality with ICC$=0.95$ exceeded the threshold for very good quality ($0.75 \leq ICC \leq 1.0$). After recordings were independently coded by the annotators, the final score for each recording fragment was determined by averaging the ratings assigned to each scale by the two annotators. We convert the 5-scale into three classification levels (low: 8.49\%, medium: 49.68\%, and high: 41.83\%) for model training and testing. 
Strictly following the annotation protocol in ~\cite{Chen2022-DyadicAP}, we annotated 16,606 five-second video clips with 1517.08 {$\pm$} 309.34 fragments from each family on average. 

\subsection{Model Training and Evaluation}

In this section, we explain the details of model training and evaluation. Once both the video clip and pose dataset (N=24,749) are shuffled, we split them into the train, valid, and test by 70\% / 20\% / 10\% ratio. For all models, we measure Top-k (k=1) accuracy with cross-entropy loss. More details about RGB frame- and skeleton-based action recognition models are described in Table \ref{table:models} and the following sections.

\subsubsection{Action Recognition Models} For RGB frame-based models, we use a base learning rate of 0.1 and it is step-wisely decayed for every 20 epochs with a total of 50 epochs. During fine-tuning, we freeze the pre-trained weight by specifying the number of layers in each backbone. The most of hyperparameters are kept the same as the default configuration provided by MMAction2 \cite{2020mmaction2}. 

\subsubsection{Skeleton-based Action Recognition Models} For skeleton-based models, we use PoseDataset as input which is extracted from NTU pose extraction \cite{duan2021revisiting}. This dataset has the format of keypoint, keypoint\_score, frame\_dir, label, img\_shape, original\_shape, and total\_frames. We set the initial learning rate is 0.1, the batch size to 128, and train models for 15 epochs with the CosineAnnealing learning rate scheduler. For the optimizer, we set the momentum to 0.9, weight decay to \num{5e-4}, and use the Nesterov momentum. The rest of the hyperparameters are kept the same as the default configuration provided by Pyskl \cite{duan2022PYSKL}.


\begin{table*}[ht]
\caption{Comparisons of detailed model configuration with state-of-the-art models. }
\centering
\begin{tabular}{p{0.1\linewidth}p{0.06\linewidth}p{0.15\linewidth}p{0.12\linewidth}p{0.11\linewidth}p{0.15\linewidth}p{0.1\linewidth}}
\hline
\centering
Models &  Year & Inputs & Data Modality & \# of Params &  Backbone & Epochs \\
\hline
\centering
TimeSformer & 2021 & $3 \times 32 \times 224 \times 224$ & RGB frame & 121.4M & TimeSformer & 15  \\
\centering

X3D & 2020 & $3 \times 16 \times 224 \times 224$ & RGB frame & 3.76M & X3D\_M & 50 \\
\centering

I3D & 2017 &  $3 \times 32 \times 224 \times 224$ & RGB frame & 28.0M & ResNet50 & 50 \\
\centering

SlowFast & 2019 & $3 \times \ 32 \times 224 \times 224$ & RGB frame & 34.6M & ResNet50-$4\times16$ & 50 \\

\hline
\centering

CTR-GCN & 2021 & $16 \times \ 2 \times 100 \times 17 \times 3$ & PoseDataset & 1.43M & CTRGCN & 15 \\
\centering

MS-G3D & 2020 &  $16 \times \ 2 \times 100 \times 17 \times 3$ &  PoseDataset &  3.17M &  MSG3D &  15 \\

\centering
ST-GCN++ &  2022 &  $16 \times \ 2 \times 100 \times 17 \times 3$ &  PoseDataset &  3.08M &  STGCN &  15 \\
\centering
ST-GCN &  2018 &  $16 \times \ 2 \times 100 \times 17 \times 3$ &  PoseDataset &  1.39M &  STGCN &   15 \\
\hline
\label{table:models}
\end{tabular}
\end{table*}

\subsection{Video Augmentation techniques}

Our dataset has imbalanced label distribution (see Section \ref{sec:dataset}) and this poses the classification task very challenging \cite{tarekegn2021review}, which may be compounded by sample size, label noise, etc. The imbalanced label distribution motivates applying video augmentation techniques. The details about each video augmentation technique will be described in the following subsections. 

  

\subsubsection{Baseline}
In Section \ref{sec:dataset}, we briefly introduce our dataset's imbalanced labels and the total number of video clips. To ensure a fair comparison with the proposed video augmentations techniques, we apply oversampling to the original dataset which duplicated 8,143 video clips from low and high joint engagement labeled video clips to make all the labels have the same ratio. In total, we prepared 24,749 video clips for Baseline and this contained 8,249 video clips per label.

\subsubsection{General Augmentations} 
To prevent the model from overfitting by "fixating" on irrelevant patterns (e.g. backgrounds), we have applied various kinds of augmentation techniques which is why we call them General Augmentations. This technique is applied to diversify the background (replace the background with RGB color, random indoor image, and blur the background), encouraging the model's robust learning by adding noise to the whole frame, randomly rotating an image, applying horizontal flipping, and lastly, giving hints of semantics in the frame by applying semantic segmentations. These different types of simple but effective techniques enabled us to supplement more features in the dataset and in total, we gathered 24,749 video clips that have all labels with the same ratio.
\label{section:datasetection}

\begin{figure*}
    \centering
    \includegraphics[width=0.80 \linewidth]{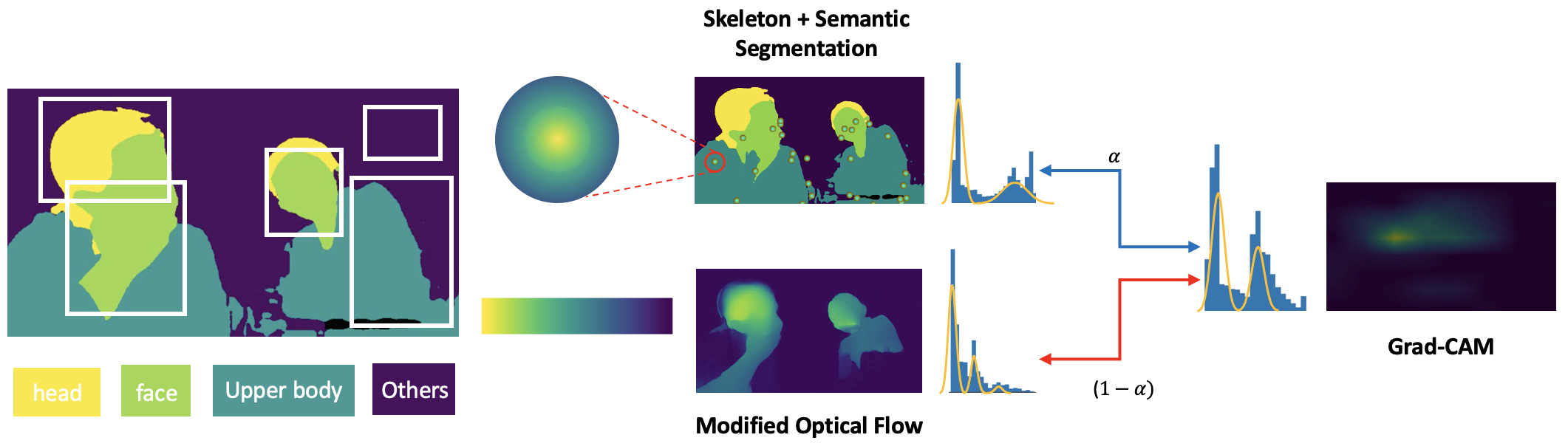}
    \caption{Pipeline for image matching metric which converts images into distributions and calculates mutual information and cross-entropy between two references (skeleton information combined with semantic segmentation, modified optical flow) and Grad-CAM.}
    \label{fig:distribution}
\end{figure*}

\subsubsection{DeepFake} 
DeepFake was applied for dyads' faces to overcome the small populations in the original dataset and also for debiasing purposes. We used SimSwap \cite{DBLP:conf/mm/ChenCNG20} for multi-person face swapping in videos. To feed a diverse set of target face images, we also utilized AI-generated face dataset (\url{https://generated.photos/faces}) which supports realistic customizations (e.g., race, gender, age, accessories, and hair type). As we can see in Table. \ref{table:overview}, this generates quite natural video clips according to its target face images. In total, we gathered 24,749 video clips as same in the General Augmentation case.

\subsubsection{Mixed} 
Finally, we also wanted to see if combining the datasets that showed performance improvement individually (see Table. \ref{table:overview}) would make even more performance improvements once combined. To do this, we randomly sampled video clips from both General Aug and DeepFake while keeping the same ratio from each dataset. So in total, we kept 24,749 video clips for Mixed. 

\subsubsection{CutOut} 
CutOut is a well-known but simple regularization technique that randomly masks out square regions of input during training (spatial prior dropout in input space) \cite{devries2017improved}. This can be used to improve the robustness and overall performance when conducting classification tasks, and in this work, CutOut is used to validate the model's representation learning without the core information in the scenes (i.e. face). To apply CutOut, we utilized the face detection module to detect the parent's and child's faces and cut out the corresponding regions, which are then replaced by black boxes. In total, we gathered 24,749 video clips by oversampling towards the largest number of labels (Mid, {$N$}=8,249) in the dataset.  

\subsection{Evaluation Metric}
\subsubsection{Gradient-weighted Class Activation Mapping}
\label{section:grad-cam}
Grad-CAM is a generalization of Class Activation Mapping (CAM) which combines the class-specific property of CAM \cite{wang2020score}. This supports the intuitive visualization of the model's attention in an image and this technique has been used in various HRI works \cite{dogan2021leveraging,kerzel2022s,gulati2021toward}. Apart from Grad-CAM's effective visualization capability, our purpose is to validate the quality of learned representations. Following the definitions in \cite{wang2020score}, Class Activation Map (CAM) and Grad-CAM are defined as follows.

\paragraph{\textbf{Definition 1. CAM}} Consider a model {$f$} with a global pooling layer {$l$} after the last convolution layer {$l-1$} and before the last fully connected layer {$l+1$}. Given a class {$c$} of interest, the CAM is defined as:
\begin{eqnarray}
L^c_{CAM} & = & ReLU(\sum_{k}\alpha_k^c A_{l-1}^k)
\end{eqnarray}

where 
\begin{eqnarray}
{\alpha_k^c} = w_{l,l+1}^c[k]
\end{eqnarray}

\paragraph{\textbf{Definition 2. Grad-CAM}} Consider a convolution layer {$l$} in a model and given a class {$c$} selected by a model, Grad-CAM is defined as: 
\begin{eqnarray}
L^c_{Grad-CAM} & = & ReLU(\sum_{k}\alpha_k^c A_l^k)
\end{eqnarray}
where 
\begin{eqnarray}
{\alpha_k^c} = GP (\frac{\partial Y^c} {\partial A_l^k}) 
\end{eqnarray}

here, {$GP(\cdot)$} denotes the Global Pooling operation.

To evaluate the quality of learned representation (Grad-CAM) which is in video format, we break this down into image frames and apply two different image-matching techniques with two references which will be explained in the following sections. 

\subsubsection{Evaluation References}
\label{section:image_matching_techniques}

In the previous section, the concept of Grad-CAM is defined and we particularly use this as a distribution learner for evaluating with two references. Here, we convert the images into distributions and calculate the mutual information and cross-entropy. The two references are 1) modified Optical Flow and 2) skeleton information combined with Semantic Segmentation. Having the two scores from each reference, we calculate the weighted average score with the weight ${\alpha}$ which has a value between 0 and 1 (See Fig. \ref{fig:distribution}).


\begin{algorithm}
    \caption{Grad-CAM evaluation algorithm}
    \label{euclid}
    \begin{algorithmic}[1]
    \State \textbf{Input:}  Images \ ${I_0}$, ${I_1}$, ${I_2}$, Model ${F(I)}$, MI($H$), CE($I$;$I$), CROP($\cdot$)
    \State \textbf{Output:} Score 
    \\
    \State Score $\gets [] $
    \While {$I_0$ \ and \ $I_1$  \ and \ $I_2$}
        \State BoundingBoxes $\gets F(I_0)$ 
        \State score  $\gets dict() $
        \For {role, part, bbox in BoundingBoxes}
             \State $I_{new0}, \ I_{new1}, \ I_{new2} \gets CROP(I_0, I_1, I_2, bbox)$
        
            \State $ mi_1   \gets MI(hist_{2d}(I_{new0}, \ I_{new1})) $  
            \State $mi_2   \gets MI(hist_{2d}(I_{new0}, \ I_{new2})) $ 
            \State $mi   \gets (\alpha \cdot mi_1 + (1-\alpha) \cdot mi_2) $
            \Comment{mutual information}
            
            \\
            
            \State $I_{p0}, \ I_{p1}, \ I_{p2} \gets Log-Softmax(I_0, I_1, I_2)$
            \State $I_{new0}, \ I_{new1}, \ I_{new2} \gets CROP(I_{p0}, I_{p1}, I_{p2}, bbox)$
            \State $ce_1  \gets CE(log-softmax(I_{new0}, I_{new1})) $
            \State $ce_2  \gets CE(log-softmax(I_{new0}, I_{new2})) $
            \State $ce  \gets (\alpha \cdot ce_1 + (1-\alpha) \cdot ce_2) $
            \Comment{cross-entropy}

            \\

            \State $score[mi][role][part] \gets mi $
            \State $score[ce][role][part] \gets ce $
        \EndFor
        \State $Score.Add(score) $
        
    \EndWhile 
    
    \State $\textbf{return} \ Score $
    
\end{algorithmic}

\end{algorithm}

The primary hypothesis supporting our framework is based on the annotation coding scheme for joint engagement (see Section \ref{sec:dataset}); subtle display (both duration and intensity) of social cues is crucial when evaluating human joint engagement (e.g., shared gaze, contingent smiling, finger pointing, etc).

\begin{table*}[htp] 
    \caption{Overview of joint engagement recognition task results from state-of-the-art end-to-end (top table) and skeleton-based (bottom table) action recognition models. Note that General Aug and DeepFake significantly outperformed the performance in all end-to-end models.}
    \label{tab:stimuli}
    \begin{tabularx}{\linewidth}{c 
                             >{}p{0.18\linewidth} 
                             c c c c}
    \hline 
    \noalign{\vskip 0.01in}  
    & \makecell{Basline} 
    & \makecell{General Aug} 
    & \makecell{DeepFake} 
    & \makecell{Mixed} 
    & \makecell{CutOut} \\ 
    
    \hline
    \noalign{\vskip 0.05in}  
    \multirow{1}{*}[2.5em]{Input} 
    & \includegraphics[height=0.65\linewidth]{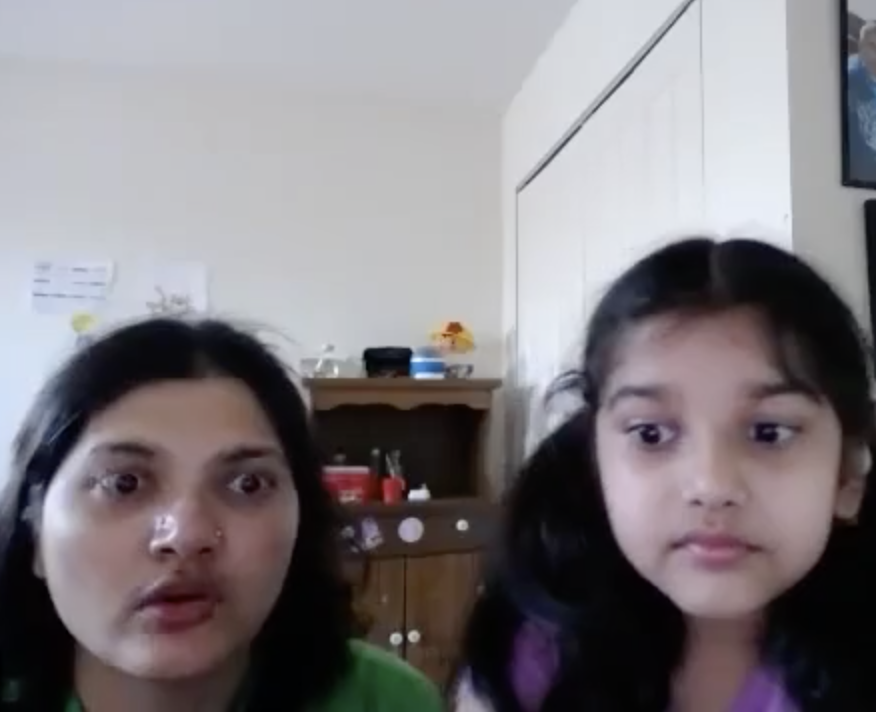}  
    &  \includegraphics[height=0.118\linewidth]{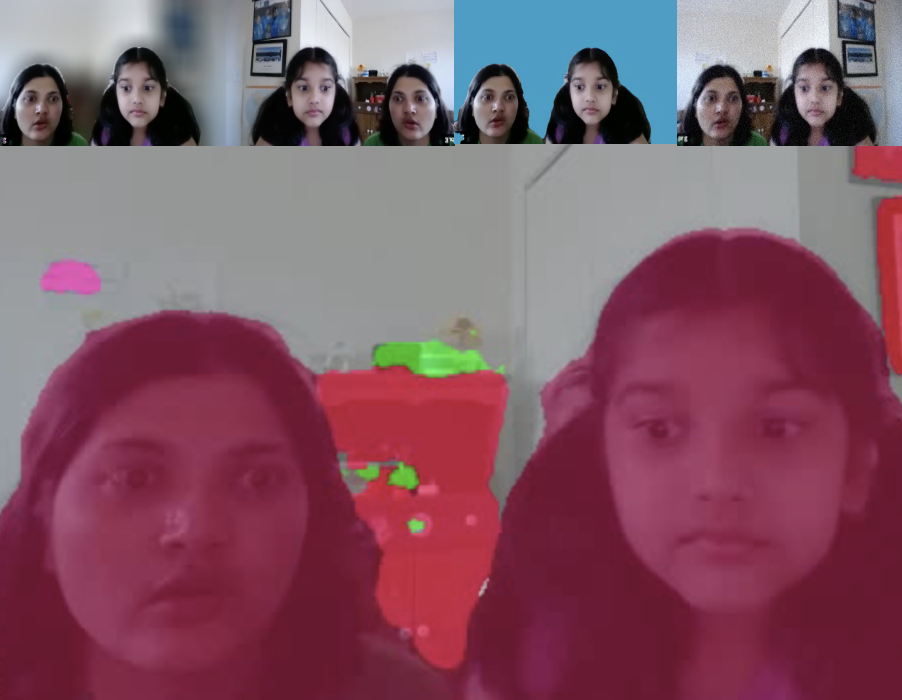} 
    & \includegraphics[height=0.118\linewidth]{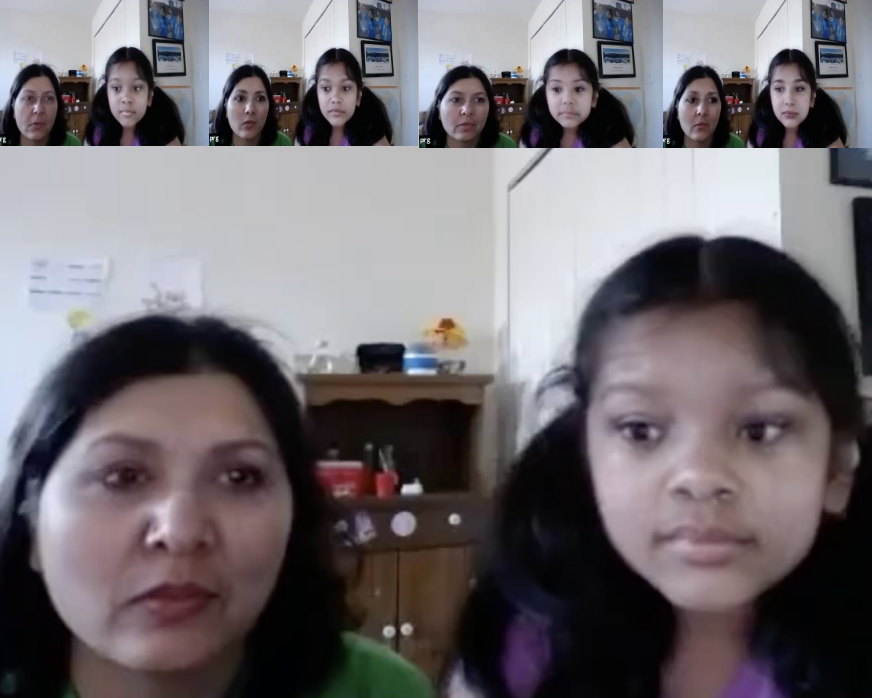} 
    & \includegraphics[height=0.118\linewidth]{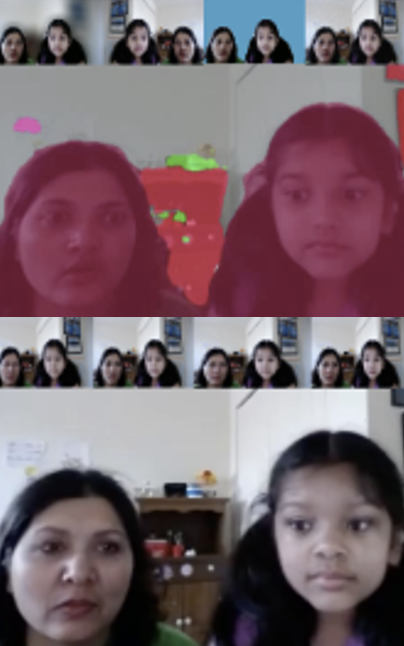} 
    & \includegraphics[height=0.118\linewidth]{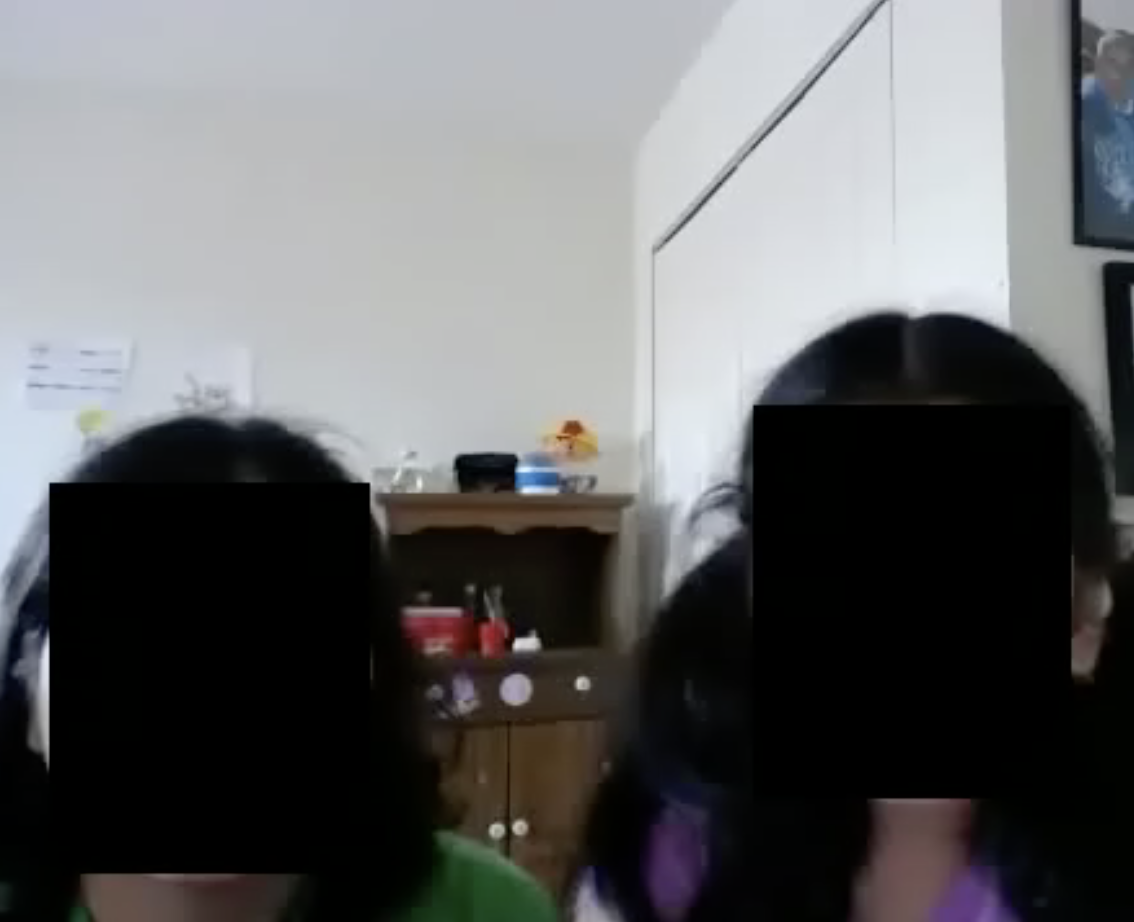} \\  
    \hline
    \noalign{\vskip 0.01in}  
    \centering
    \multirow{1}{*}{TimeSformer Cls (\%)} 
    \centering 
    & 62.2 \ (+0.0) 
    \centering
    & 66.7 \textcolor{forestgreen}{(+4.5)}  
    \centering
    &  \textbf{70.1 \textcolor{forestgreen}{(+7.9)}} 
    \centering
    &  63.0 \textcolor{forestgreen}{(+0.8)}  
    \centering
    &  60.9 \textcolor{red}{(-1.3)}    \\

    
    \noalign{\vskip 0.01in}  
    \centering
    \multirow{1}{*}{I3D Cls (\%)} 
        \centering
        & 60.1 \ (+0.0) 
        \centering
        & 63.4 \textcolor{forestgreen}{(+3.3)}   
        \centering
        &  \textbf{66.9 \textcolor{forestgreen}{(+6.8)}} \centering 
        & 60.7 \textcolor{forestgreen}{(+0.6)}
        \centering
        & 55.8 \textcolor{red}{(-4.3)}    \\ 
    
    \noalign{\vskip 0.01in}  
    \centering
    \multirow{1}{*}{SlowFast Cls (\%)} 
        \centering
        & 61.2 \ (+0.0) 
        \centering
        &  \textbf{63.8 \textcolor{forestgreen}{(+2.6)}} 
        \centering
        &  62.2 \textcolor{forestgreen}{(+1.0)}  
        \centering
        &   58.4 \textcolor{red}{(-2.8)}
        \centering
        &  50.0 \textcolor{red}{(-10.2)}  \\ 
    \hline
    \noalign{\vskip 0.05in}  
    
    \multirow{1}{*}[2em]{Input} 
    \centering
        & 
        \centering
        &  
        \centering
        & \includegraphics[height=0.08\linewidth]{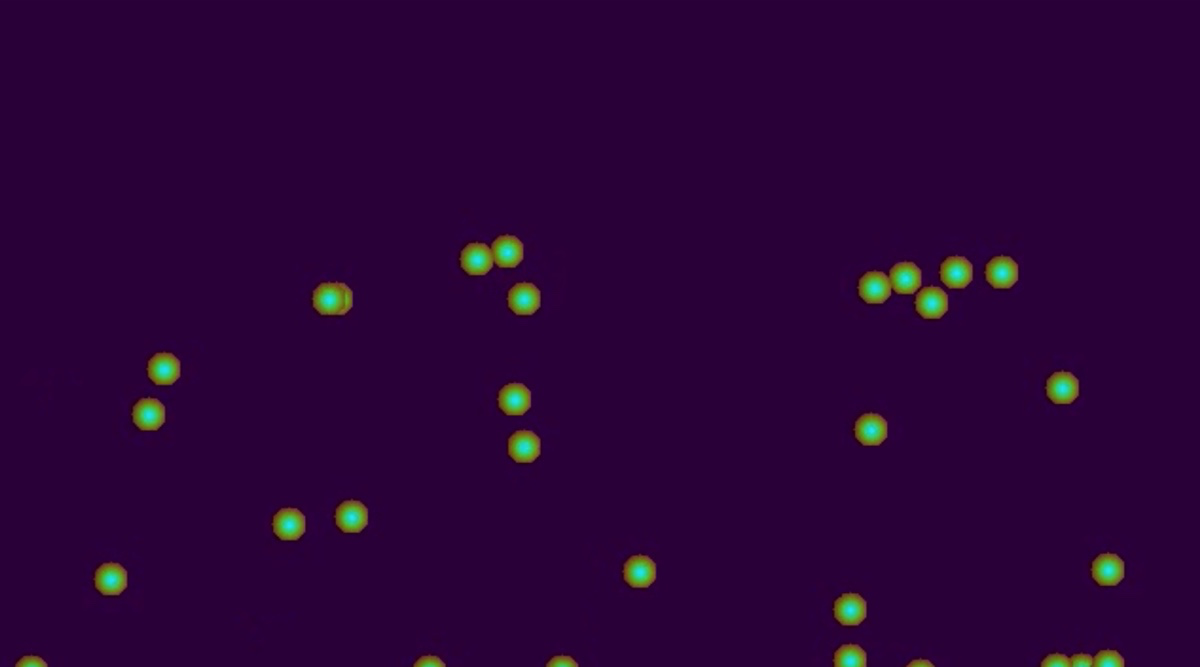} 
        \centering
        & 
        \centering
        & \\
    \hline
    
    \noalign{\vskip 0.01in}  
    \centering
    \multirow{1}{*}{CTR-GCN Cls (\%)} 
        & &  & \textbf{64.3} & & \\ 
    
    \noalign{\vskip 0.01in}  
    \multirow{1}{*}{MS-G3D Cls (\%)} 
        & &  & 63.3 & & \\ 
    
    \noalign{\vskip 0.01in}  
    \multirow{1}{*}{ST-GCN++ Cls (\%)} 
        & &  & 64.1 & & \\ 
    
    \noalign{\vskip 0.01in}  
    \multirow{1}{*}{ST-GCN Cls (\%)} 
        & &  & 59.4 & & \\
    \hline
      
  \end{tabularx}
\end{table*}

Accordingly, to recognize the subtle motion changes, we apply Optical Flow from \cite{2021mmflow}. After that, we modify the original Optical Flow which displays the color by their orientations, but instead, we discard this orientation-based colormap and follow the colormap used in Grad-CAM to focus on the motion changes itself (See Fig. \ref{fig:framework}). Also, to ensure the learned representation is focusing on the proper regions, we considered skeleton information combined Semantic Segmentation as the other reference. We first extract the skeleton information by applying the pose extractor described in \cite{duan2021revisiting} and combine this with the Semantic Segmentation results \cite{li2020self}. For Semantic Segmentation, we specify each body segment with pre-defined colors (see Fig. \ref{fig:distribution}), and when combining, the Gaussian heatmap is centered on each of the body poses (see Fig. \ref{fig:distribution}). This is based on the insights from our annotation process where we evaluate the dyad's joint engagement level by focusing on social touch, body closeness, heading angle, and smiling. Since our dataset mostly contains the face and upper parts of the body, we put more importance on the head and the upper body parts rather than the bottom parts of the body.    

\subsubsection{Image Matching Techniques}
\label{section:image_matching_techniques2}

First, we adopt mutual information, a dimensionless quantity metric that measures the mutual dependence between two variables. The metric is high when the attention map signal is highly concentrated in a few histogram bins, and low when the signal is spread across many bins. Mutual information is defined as:

\begin{eqnarray}
I(X;Y) = \sum_{x \in X} \sum_{y \in Y} p(x,y)log(\frac{p(x,y)}{p(x)p(y)})
\end{eqnarray}

Here, we convert the image into a distribution by flattening the image arrays and then compute the bi-dimensional histogram of two image array samples. The second metric is cross-entropy, which comes from the Kullback-Leibler divergence. This is a widely used metric for calculating the difference between two distributions, and this is defined as:
\begin{eqnarray}
H(p,q) = -E_{p}[log \ {q}]
\end{eqnarray}

where $ E_{p}[\cdot]$ \ is the expected value operator with respect to the distribution $p$. 
Here, we first normalize the pixel values in images and then pass this through log-softmax to convert images into distributions. Then we apply cross-entropy as explained in Equation 6. Given that we have all the bounding boxes for each parent's and child's face and body, we could separate these values based on their bounding box coordinates (See Fig. \ref{fig:framework}). The overall process is summarized in Algorithm 1. where the model {$F(I)$} is the face, age, and gender detector, {$MI(H)$} is mutual information function, {$CE(I;I)$} is a cross-entropy function, and {$CROP(\cdot)$} is a image crop function.

\section{Experiment and Results}
In this section, we conduct experiments to evaluate the effectiveness of the proposed video augmentation techniques; General Aug, DeepFake, Mixed, and CutOut to see their capability to improve joint engagement recognition on state-of-the-art action recognition models. Also, we compare the joint engagement classification performance between RGB frame-based and skeleton-based models to see the effect of different inputs in this task. For the implementation, we utilized MMAction2 and Pyskl, an open-source toolbox for video understanding based on PyTorch and all of the models were trained on 8 NVIDIA 1080Ti GPUs. Source code and pre-trained models are available at [Hiddenforannonymoussubmission].


\begin{figure*}
    \centering
    \includegraphics[width=0.8\linewidth]{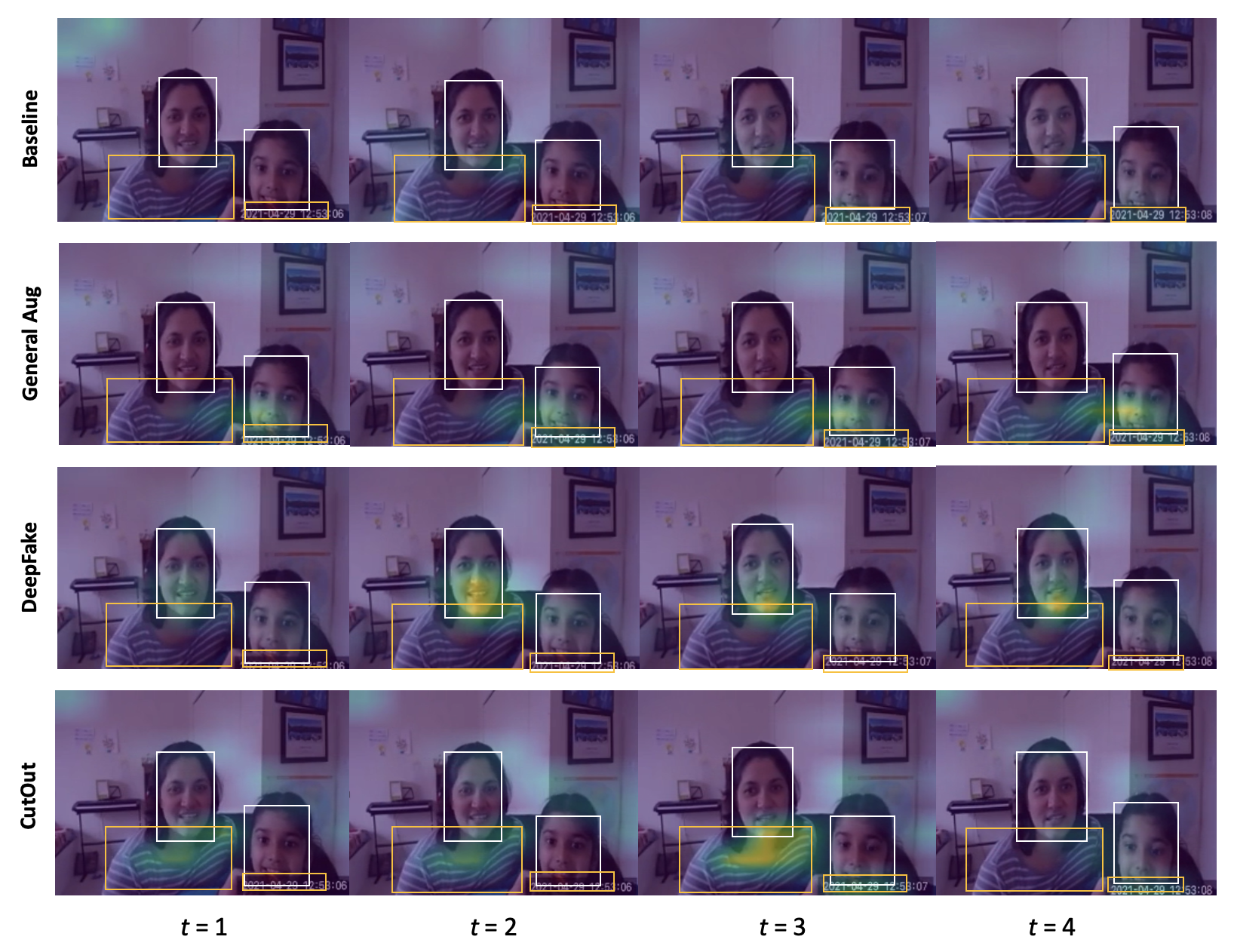}
    \caption{Grad-CAM visualization of a sample test video clip generated from fine-tuned I3D with different video augmentation techniques (Mixed is excluded here since it was only used for the performance comparison).}
    \label{fig:finalplot}
\end{figure*}

\subsection{Joint Engagement Recognition Evaluation}

As shown in Table. \ref{table:overview}, applying General Aug and DeepFake consistently outperformed the baselines in all end-to-end models. Particularly, TimeSformer and X3D lead the accuracy by up to \textbf{15.7\%} and \textbf{14.5\%} respectively. On the other hand, using CutOut did not improve the performance compared to the baseline performance for all end-to-end models. However, since the performance did not degrade compared to the baseline in CutOut, we interpret this as the model would try to focus more on the behavioral aspects rather than utilizing the features from the face. Also, in Mixed, the performance increases except for the case of SlowFast. This shows all of the end-to-end models here, lack diverse features to generalize compared to the case in General Aug and DeepFake. These results give us an insight into the generalization of our video augmentation techniques for joint engagement recognition. 

The results from the skeleton-based models (see Table. \ref{table:overview}) however, show lower top-1 accuracy (max acc: 64.3\%) in all cases compared to the baseline in end-to-end models. These graph convolution network-based models attempt to find Spatio-temporal patterns from the human skeleton information. However, unlike human action recognition tasks, joint engagement recognition requires the model to catch the affective state (e.g., facial features) of people which RGB frames particularly include and this makes skeleton-based models challenging to learn joint engagement without the core information.

\subsection{Visualizing Interaction Representations}
To have a deeper understanding of what the models have learned, we use Grad-CAM to visualize the Spatio-temporal regions that contribute the most to classify into certain joint engagement classes on the dataset (see Fig. \ref{fig:finalplot}). We observe that the learned representations focus on either small regions in the "face and body" or get distracted by the backgrounds. For example, in Fig. \ref{fig:finalplot}, both the Grad-CAM in the General Aug and DeepFake groups match well with either parent's or child's face regions, but in the Baseline, the heatmap is not actively paying attention to dyads but looking at the corner in the scene. This implies that training on a diversified set of populations or backgrounds encourages the model to be more robust about the change of backgrounds, clothing, etc, and focus more on dyads’ faces and bodies. Also, in CutOut, the model did not fully focus on the core parts of the scene but got distracted by the backgrounds. This is because it could not refer to the face part which includes the core information during the training.


\section{Discussion and Conclusion}
The performance and visualization of the state-of-the-art end-to-end video classification models for recognizing joint engagement demonstrated their potential to recognize complex human-human joint affective states with limited training data. The fine-tuned end-to-end models initially pre-trained for general video understanding (e.g., SlowFast, I3D) performed more effectively on joint engagement recognition than the models trained on human skeleton features (e.g., CTR-GCN, ST-GCN). The video augmentation techniques enhanced the model's performance even further. The visualization of the learned representations in the end-to-end deep learning models revealed their sensitivity to subtle social cues indicative of parent-child interaction. Altogether, these findings and insights indicate that end-to-end models were able to learn the representation of parent-child joint engagement in an interpretable manner. 

Due to their superior performance and interpretability, end-to-end models pre-trained on action recognition tasks provide new opportunities for developing autonomous robots for multi-person human-robot interaction. When skeleton-based models are used for real-time affect perception in the wild, typically multiple layers of models are necessary for human identification, human skeleton feature extraction, and affect prediction (e.g.,~\cite{Alghowinem2021-BodyGH}). In contrast to skeleton-based models, end-to-end models accept image frames as input and generate an affect prediction as output. After being trained, they can be utilized to make predictions in real-time without requiring excessive computational resources. The real-time version of the SlowFast model, for instance, can function with minimal computational resources~\cite{wei2022efficient}. Similarly, the visualization method we proposed provides a means of demonstrating how recognition models extract crucial information from videos for the learning task. This interpretable visualization could be used to support diverse methods of robot learning in real-time from human input or human teachers, such as the "social scaffolding for exploration" method and socially guided machine learning~\cite{thomaz2016computational}.

We acknowledge that the relatively small population size of the dyads may limit the applicability of the proposed framework for multi-person affect recognition. However, the small size of the dataset motivates the use of a framework that can leverage various data augmentation techniques and deep learning models pre-trained on larger data corpora for other similar tasks. Indeed, the excellent performance of our proposed framework on the small dataset demonstrates its potential benefits for real-world multi-person HRI, as the real-world interaction datasets used to train a robot's perception model may not be as large as datasets parsed from the Internet or generated by simulation. We also acknowledge that the analysis in the visualization of the model's learned representation does not pinpoint the specific social cues and behavioral characteristics that guide the model's recognition of joint engagement. Our work only aims to demonstrate that the visualization of the model's learned representation can reveal semantically and interpretable insights that can be valuable to humans and can potentially allow humans to correct the model, challenging the widely held belief that end-to-end models have limited interpretability in comparison to skeleton-based models.

In the future, we plan to use additional parent-child interaction datasets to investigate how to generalize this framework across datasets and to quantify how various social and behavioral cues contribute to the learned representations of end-to-end models. Additionally, our proposed framework is extensible in multiple ways. First, the proposed data augmentation and visualization can be applied to multiple data modalities, such as audio and video, to jointly learn the joint engagement. Utilizing multiple modalities in the dataset would increase the applicability of multi-person affect models to challenging real-world situations, such as missing data in one modality, and further enable the affect model to learn more holistic and human-like representations of joint engagement.

Our proposed framework can also be expanded to account for individual differences in affect across dyads by adding a deep neural network layer as the final layer trained on individual human groups. It has been empirically demonstrated that personalized and culture-sensitive affect models outperforms one-size-fits-all generic models when recognizing affect of individuals (e.g., ~\cite{Rudovic2018-CDL}, ~\cite{Rudovic2018-PML}). Thus, personalized or culture-sensitive joint engagement models may further enhance the performance of model predictions in the multi-person context. 
 
In conclusion, this work serves as the first step toward fully unlocking the potential of state-of-the-art end-to-end video understanding models pre-trained on large public datasets and augmented with data augmentation and visualization techniques for robot's affect recognition in the multi-person human-robot interaction in the wild.

\bibliographystyle{ACM-Reference-Format} 
\bibliography{sample}

\end{document}